\documentclass[letterpaper, 10 pt, journal, twoside]{IEEEtran} 
\IEEEoverridecommandlockouts   
\usepackage{xcolor}
\usepackage{hyperref}
\usepackage{graphicx}
\usepackage{textcomp}
\usepackage{float}
\usepackage{pgfplots}
\usepackage{array}
\usepackage{amsmath}
\usepackage{cleveref}
\usepackage{multirow}
\usepackage{comment}
\usepackage{enumitem}
\usepackage{cite}
\newcolumntype{P}[1]{>{\centering\arraybackslash}p{#1}}
\newcolumntype{M}[1]{>{\centering\arraybackslash}m{#1}}

\usepackage{amsthm}
\theoremstyle{definition}

\newlist{conditionenum}{enumerate}{1}
\setlist[conditionenum,1]{
    label={\textbf{C\arabic*}},
    ref=\arabic*,
    labelindent=1.5em,
    labelsep*=0.5em,
    leftmargin=*
}
\creflabelformat{conditionenumi}{#2\textup{\textbf{C#1}}#3}
\crefname{conditionenumi}{condition}{conditions}
\Crefname{conditionenumi}{Condition}{Conditions}

\title{A Methodology for Approaching the Integration of Complex Robotics Systems Illustrated through a Bi-manual Manipulation Case-Study}

\author{Pavlos Triantafyllou*, Rafael Afonso Rodrigues*, Sirapoab Chaikunsaeng*, Diogo Almeida, Graham Deacon, Jelizaveta Konstantinova, Giuseppe Cotugno\dag%
\thanks{This work has received funding from the European Union's H2020 programme under grant agreement No. 643950, project SecondHands.}
\thanks{All authors are with the Robotics Research Team, Ocado Technology, Hatfield, UK.} 
\thanks{* Authors contributed equally. \dag Corresponding author.}
}

\begin{document}

\maketitle

\begin{abstract}
The multidisciplinarity of robotics creates a need for robust integration methodologies that can facilitate the adoption of state-of-the-art research components in an industrial application.
Unfortunately, there are no clear, community-accepted guidelines or standards that define the integration of such components in a  single robotic system.
In this paper, we propose a methodology that assesses the software components of a candidate system on the basis of the effort required to integrate them and the impact their integration will have on a target system. We demonstrate how this methodology can be applied using an industrial tool packing system as an example. The system integrates 
a wide range of both in-house and third-party research outputs and software components. We prove the effectiveness of our approach by evaluating system performance with an experimental benchmark that assesses the robustness, reliability and operational speed of the system for the given packing task.
We also demonstrate how our methodology can be used to predict the amount of integration time required for a component.
The proposed integration methodology can be applied to any robotic system to facilitate its transition from the research to an industrial environment.

\end{abstract}

\section{Introduction}

In industrial contexts, classical applications of robotics rely on rigid assumptions about the robot's physical environment to simplify system development.
However, recent advancements in robotics research have led to an increasing number of systems working in challenging real-world conditions.
For instance, complex robotic systems were developed for  applications such as autonomous driving \cite{autonomous_driving_2011}, warehouse automation \cite{warehouse_automation2018} and disaster response \cite{search_and_rescue2012}.
Common to all these systems from different application areas is that they simultaneously address complex research topics, such as motion planning, perception and control, in a context of significant environmental uncertainty.
This multidisciplinarity, combined with the novelty of each component, emphasizes the need for robust integration methodologies.

\begin{figure}[t]
    \centering
    \includegraphics[width=0.45\textwidth]{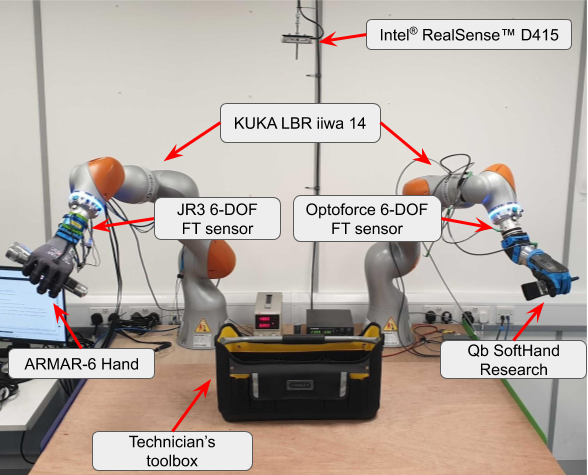}
    \caption{The dual-arm tool-packing system at the OCADO Robotics Research Laboratory developed using the methodology presented in this paper. It consists of two KUKA LBR iiwa 14 R820 arms equipped with different combinations of FT sensors and robotic hands. One arm has a JR3 FT sensor and an ARMAR-6 Hand, whereas the other has an Optoforce FT sensor and a qb SoftHand. An Intel\textregistered{} RealSense\texttrademark{} D415 depth camera is used to perceive the scene.}
    \label{fig:bimanual_system}
\end{figure}

The integration of robotic systems is more commonly discussed in the context of robotics competitions, e.g. \cite{darpa, Yu2016, Schwarz2017, eppner2018, hofmann2019}, rather than for real-world industrial applications. 
These works focus on the architecture of the system, rather than on the adopted integration solutions.
The consequence of this approach is that knowledge and experience on integrating complex robotic systems is not shared with the robotics community. 
As such, the inclusion of third-party contributions into a new robotic system remains a challenging task \cite{cervera2019}.
Some exceptions to this trend include \cite{Hernandez2016, Morrison2018}, which detail ROS-based component approaches adopted in the winning systems of the Amazon Picking/Robotics Challenge in 2016 and 2017, respectively.
A subsequent postmortem of the developed system in \cite{Hernandez2016}, further emphasizes the need for systematic approaches to integration, and analyses the system from a functional perspective \cite{Corbato2020}.
These works, however, do not provide a systematic approach to estimate the integration effort of arbitrary components.
Moreover, while ROS is a suitable framework for component-integration, it can be impractical when the system integrator does not have the time or the ability to adapt a system component to ROS. 
It can be observed that the integration approaches are not broadly shared by the robotic community. Therefore, methodologies cannot be compared and it is difficult to learn from mistakes and build widely accepted good practices. 
This ultimately makes the transfer of results from academia to industry more difficult, as methods which perform well in lab conditions turn out to be under-performing or unusable due to technical and integration limitations not accounted for or documented in the original research paper.

To address this issue, we propose a holistic integration methodology that assesses candidate system components on an effort-impact basis. This helps in bridging the gap between academic research and industry. This paper presents the practical requirements for the integration of an industrial system and can be considered when exploiting the research contributions towards specific use cases. Our approach indicates how to technically integrate different components in a working system and it is designed to be employed alongside a development methodology, like Agile Scrum \cite{lei2017}, which indicates how teams should organise their work. Our work is part of the EU H2020 SecondHands\footnote{SecondHands project website: \url{http://secondhands.eu/}} project, which aims at developing a collaborative robot able to assist a maintenance technician in several tasks such as packing maintenance tools in a toolbox. The project was inspired by the industrial use case of Ocado, which aims to transform the online grocery business through cutting-edge technology and innovation. The resulting tool packing system (see Fig. \ref{fig:bimanual_system}) is an example of complex robotic system successfully applied to a real-world industrial use case thanks to a robust and comprehensive integration methodology, illustrated in this paper. The tool packing system presented in this paper is used to illustrate in practice how our methodology can be applied to a real-world complex robotic system. 

\begin{figure*}
	\centering
	\includegraphics[width=0.98\textwidth]{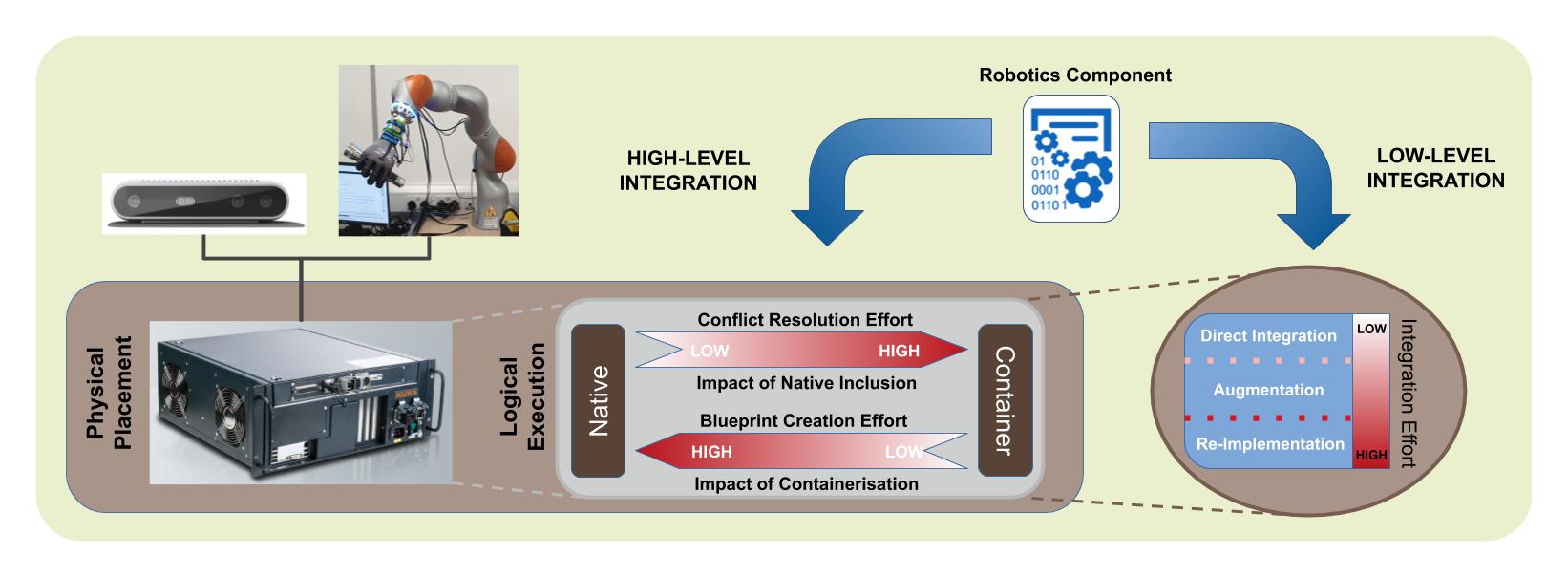}
	\caption{This diagram illustrates the integration methodology for the proposed system. The various integration aspects are managed independently.
	\textit{High-level integration}: a new robotics component is first assigned to a specific machine (Physical Placement). Then a decision on how to make it compatible with the system's \textit{development environment} (Logical Execution) is made based on an effort-impact assessment. \textit{Low-level integration}: the same component is integrated or an alternative component is developed from scratch. The proposed effort-based metric is used as a guideline for deciding on an integration approach (Section \ref{low_level_integration}).
	Color gradients represent the level of effort or impact, from low (light red) to high (dark red).}
	\label{fig:integration_diagram}
\end{figure*}

The main contributions of this paper are summarised as follows:
\begin{enumerate}
    \item An integration methodology for the development of complex robotic systems, with an emphasis on the estimated integration effort of any given component;
    \item A case study which employs our methodology for the development and integration of a dual-arm packing system;
    \item An assessment protocol and a benchmark to experimentally validate the proposed methodology aiming at a good system performance. 
\end{enumerate}

Our methodology is a novel approach to robotics integration which is addressing the problem of integration systematically. Moreover, with this methodology we aim to bridge the gap between fundamental and industrial research, facilitating easier adoption and integration of research outputs. To achieve this it relies on concepts which are well known in software development, such as virtualisation, and applies it to the domain of robotics software integration. 
The rest of this article is structured as follows.
We present the integration methodology in Section \ref{systems_integration}.
Our dual-arm packing system is described in Section \ref{dual_arm_system}.
In Section \ref{case_study}, the system is used as the basis for a case study where we describe the application of the integration methodology.
An experimental evaluation of our methodology
and the results are included in Section \ref{experiments}, and Section \ref{discussion} discussed the results and the implications of our approach. Section \ref{conclusion} concludes the paper. 
\section{SYSTEMS INTEGRATION METHODOLOGY} \label{systems_integration}

\begin{figure*}[t!]
	\centering
	\includegraphics[width=0.7\textwidth]{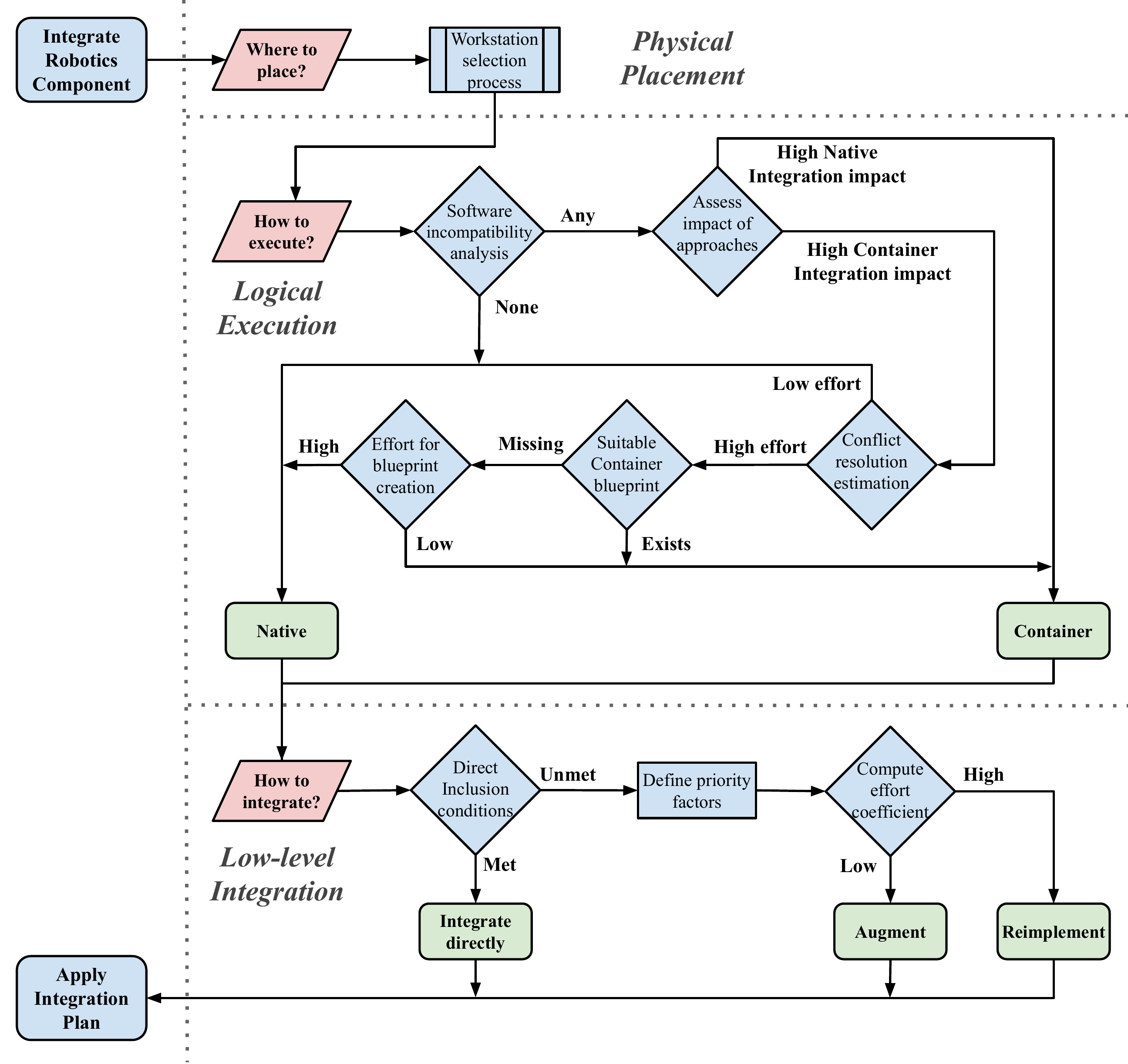}
	\caption{This flowchart depicts the decision process of how to integrate robotics software components using the proposed methodology.}
	\label{fig:integration_flowchart}
\end{figure*}

In this section, we propose a methodology that can be used to successfully guide the development and integration of complex robotics systems. This methodology can be introduced at any stage of development and does not need to be adopted from the beginning. An overview of this methodology is illustrated in Fig. \ref{fig:integration_diagram}. In order to distinguish between different aspects of integration, we define the following integration levels:
\begin{itemize}
    \item \textit{High-level integration} represents how a component is integrated in a \textit{compatible way} with the {development environment} of the system. In the context of this paper, a development environment consists of the hardware (processor, memory, etc.) and software (operating system, drivers, etc.) specification of the machine(s) on which the system's code is executed.
    \item \textit{Low-level integration} represents how a component is integrated in order to ensure \textit{interoperation}  with the rest of the system and function within the \textit{quality and performance levels} required by the end-user.
\end{itemize}
The integration methodology workflow is summarised in Fig. \ref{fig:integration_flowchart}, which illustrates the decisions to take for the two integration levels.

When developing the tool packing system, we paired our integration methodology with the Agile Scrum development methodology \cite{dybaa2008}, however other development methodologies, such as Kanban \cite{ahmad2013} or even Waterfall, could also be employed. The rest of this Section is organized as follows. First, we describe two different approaches for high-level integration and present qualitative assessment criteria to select the appropriate integration (Section \ref{high_level_integration}). Secondly, we discuss  various aspects of low-level integration and propose a metric that quantifies the amount of effort required for low-level integration of a certain component (Section \ref{low_level_integration}).

\begin{table*}[t!]\label{integration_effort_table}
    \caption{Examples of development efforts required to meet  low-level integration conditions.}
    \label{integration_effort}
    \begin{center}
        \begin{tabular}{|M{0.12\linewidth}||M{0.25\linewidth}|M{0.25\linewidth}|M{0.25\linewidth}|}
\hline
\textbf{Condition (No.)} & \textbf{Low Effort} & \textbf{Medium Effort} & \textbf{High effort} \\
\hline
Interface \textbf{(C\ref{req_interface})} & An API exists and can be used with minor modifications & An API exists, but requires significant modifications to be used & There is no API, so one has to be implemented from scratch  \\
\hline
Syntax \textbf{(C\ref{req_syntax})} & The component is not written in the same language as the target system, but wrappers for this language are provided & The component is not written in the same language as the target system and wrappers for this language are not provided &
The component is written in a framework which is not directly compatible with the target system
\\
\hline
Performance \textbf{(C\ref{req_performance})} & Performance optimization can be achieved with moderate software modifications & Performance optimization requires a hardware update, but can be achieved without modifying the component's logic & Complete rewriting of the component's logic is required for it to be efficient enough\\
\hline
Code quality \textbf{(C\ref{req_quality})} & Code is tested extensively but not thoroughly documented or vice versa & Code is partially documented and/or tested & Code follows poor coding practices, lacks documentation and test coverage
\\
\hline
Support \textbf{(C\ref{req_support})} & The component is actively maintained but there is no official support & The component is unofficially maintained and supported & The component has been deprecated and no support exists
\\
% \hline
% Licensing \textbf{(C\ref{req_license})} & - & Licensing requires software modifications to be made public & Licensing prohibits commercial usage
% \\
\hline
\end{tabular}
    \end{center}
\end{table*}

\subsection{High-level Integration}
\label{high_level_integration}
High-level integration of components in distributed systems comprises the analysis of the software requirements components have in order to determine their \textit{logical execution}.
For instance, a certain component might require a specific set of drivers or libraries to be installed in the system's development environment.
This aspect is researched less, particularly in robotics, and appears in contrast to \textit{physical placement}, which identifies the specific hardware (computer) where a component runs.
We consider the following types of logical execution: 

\begin{itemize}
    \item \textit{Native inclusion} is the execution of a system component directly in a development environment without mediation.
    \item \textit{Containerisation} is the execution of a system component within a virtual environment. The virtual environment mediates every interaction of the component with the development environment, and the component is isolated from the rest of the software running on the machine. 
\end{itemize}

Until recently, native inclusion was the only approach for integrating robotics software \cite{docker17}. The underlying concept is straightforward and, in general, easy to apply. But this simplicity is not representative of the integration efforts required when incompatibilities arise between a new component and the existing system or development environment. For instance, if system components require incompatible drivers or operating systems, then the only way to include them natively is to use more computers. Whereas, several components might require different versions of the same library which will have to be installed concurrently to avoid software conflicts. Solving these software conflicts is challenging and can lead to functional and structural impacts on the system. Functional impacts correspond to restrictions in system functionality. For example, a dependency on specific drivers might limit the functions a system can use. Structural impacts affect the ability to maintain and expand the system in the future. For instance, manual compilation of software from source does not scale well and disables the use of tools for updating and maintaining software automatically, negatively affecting the system robustness.

Containers (Docker\footnote{https://www.docker.com/}, LXC\footnote{Linux Containers, https://linuxcontainers.org/}) aim to solve the aforementioned incompatibility problems by providing each component with a dedicated software development environment (operating system, drivers, libraries, etc.). This makes the system very modular and easy to deploy, and avoids the negative functional and structural effects present in native inclusion. Furthermore, isolated execution of components facilitates containing and mitigating security breaches and other vulnerabilities that may originate from a certain component. However, containerisation has a setup cost, as the complete software development environment of a component must be described in a template format, namely the \textit{container blueprint}. Creating and testing this blueprint can be a tedious process. Moreover, as containers are not widely used in robotics yet, there is also a learning and adaptation cost associated with their use. Another factor to consider when systems have limited computational resources is the impact on performance resulting from the mediation between virtual and development environments \cite{felter2015}.

Overall, the choice between {native inclusion} and {containerisation} of a component is influenced by the integration \textit{effort} required and the \textit{impact} that an integration approach has on the system. The correct estimation of those two factors by the development team is therefore essential for estimating the expected time required for the high-level integration. A mapping between integration effort and time per factor is presented in Table \ref{time_table}. The numbers are derived from the analysis of our experience in the integration of complex robotic systems, including the tool packing system described in this paper. 

\begin{table}[t!]
    \caption{Mapping between integration effort and time for high-level integration.}
    \label{time_table}
    \begin{center}
        \begin{tabular}{|M{0.2\linewidth}||M{0.15\linewidth}|M{0.15\linewidth}|M{0.15\linewidth}|}
            \hline
            &
            \textbf{Low effort} & 
            \textbf{Medium effort} & 
            \textbf{High effort} 
            \\
            \hline
            \textbf{Integration time} &
            0-2 person days & 
            3-6 person days & 
            $>$6 person days \\
            \hline
        \end{tabular}
    \end{center}
\end{table}

\subsection{Low-level Integration}
\label{low_level_integration}
Low-level integration defines how a component interacts with the rest of the system and functions according to defined quality and performance levels. To estimate the effort required for the integration and subsequent use of a component, system developers should assess the  development states of the components both in isolation and in combination with the target system. Specifically, we propose that a component should satisfy the following conditions:

\begin{conditionenum}
	\item To offer software interfaces (APIs) that can be used without modifications by the target system; 
    \label{req_interface}
	\item To be written in the same programming language and/or framework as the target system, or to provide ways to interoperate with the system; \label{req_syntax}
	\item To perform efficiently in terms of execution time and resource consumption; \label{req_performance}
	\item To follow the well-established coding practices, such as thorough documentation and testing;
	\label{req_quality}
	\item To be actively maintained. \label{req_support}
\end{conditionenum}

The satisfaction of the above conditions guarantees that the integration of a component in the overall system does not negatively affect the system's interoperability (\cref{req_syntax,req_interface}), performance (\cref{req_performance}), maintainability (\cref{req_syntax,req_quality,req_support}) and transparency (\cref{req_quality}). If a component satisfies all these conditions for the target system, then it can be \textit{integrated directly} without any further modification. Otherwise, a varying amount of integration effort is required, depending on the degree to which each condition is met (see Table \ref{integration_effort}). In this case the following approaches can be considered:
\begin{itemize}
    \item \textit{Augmentation}: to implement the necessary extensions and/or improvements for a component to be integrated in the system.
    \item \textit{Re-Implementation}: to develop an alternative version of the component that satisfies the system's interoperability, performance and quality requirements.
\end{itemize}

To decide between the above two approaches of  \textit{augmentation} and \textit{re-implementation}, one must quantify the amount of effort required to adapt a component to the target system's requirements. Therefore, we define the following scoring system to assess a single component on each of the aforementioned conditions: condition satisfied (0), low effort (1), medium effort (2), high effort (3). Based on this, we derive an effort coefficient $e_f \in [0,1]$ as follows:

\begin{equation}\label{integration_equation}
e_f = \frac{C_{total}}{C_{max}} = \frac{\sum\limits_{i\in[1,5]}{p_{C_i}*E_{C_i}}}{\sum\limits_{i\in[1,5]}{p_{C_i}*3}},
\end{equation}

where $E_{C_i} \in [0,3]$ represents the effort score and
$p_{C_i} \in [0,1]$ is the \textit{priority factor} for the $i^{th}$ condition. Priority factors help define which of the conditions are more relevant during the integration of the components in the target robotic system.
$C_{total}$ represents the total amount of integration effort required for a given component as the sum of the effort scores weighted by the corresponding priority factors. 
$C_{max}$ corresponds to the worst-case scenario where all conditions require a high amount of effort to be satisfied; this corresponds to an effort score $E_{C_i}=3$ per condition. 

High $e_f$ values mean that a significant amount of effort and time is required to make a component compliant with the system's performance, interoperability and quality requirements. Therefore, the higher the effort coefficient is, the higher the perceived value of a component must be to justify the amount of integration effort required. We model the relationship between $e_f$ and integration time $t_i$ in person days as follows:
\begin{equation}\label{integration_time_equation}
t_i = 10 * e_f.
\end{equation}
Conditions \textbf{C1} (Interface), \textbf{C2} (Syntax) and \textbf{C3} (Performance)
affect integration time owing to the amount of code that needs to be written to meet them, whereas conditions \textbf{C4} (Code quality) and \textbf{C5} (Support) impact the amount of time needed to understand and test the code.
\section{DUAL-ARM TOOL-PACKING SYSTEM \label{dual_arm_system}}

In this section, we present a dual-arm manipulation system that we developed using the methodology presented in Section \ref{systems_integration}. First,
we describe the manipulation task in Section \ref{task_description}.
Then, Section \ref{system_setup} details the system setup. Finally, we present an overview of the system's functionality in Section \ref{system_overview}.

\subsection{Task description} \label{task_description}
\begin{figure}[t]
    \centering
    \includegraphics[width=0.45\textwidth]{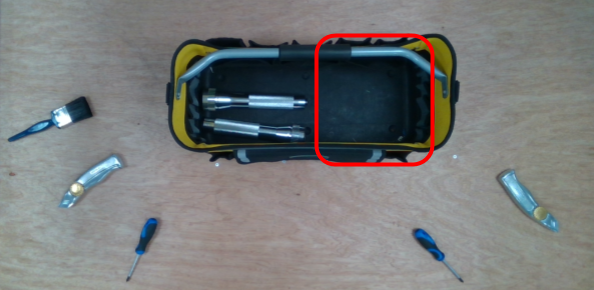}
    \caption{Example setting for the tool-packing task with the proposed dual-arm system. The red rectangle denotes the area of the toolbox where the next tool must be placed so that space in the toolbox becomes  uniformly occupied.}
    \label{fig:task_setting}
\end{figure}

 Figure \ref{fig:task_setting} represents the example task for the dual-arm robotic manipulation system (Fig. \ref{fig:bimanual_system}. A set of maintenance  tools are positioned on the flat working surface.  As part of the task, the system must autonomously pack all the tools in the toolbox. Each robot arm must grasp a single item at a time, and the area inside the toolbox should be occupied uniformly. Therefore, the initial placement target identified on Figure \ref{fig:task_setting} lies within the empty toolbox area marked by the red rectangle.

Therefore, we define the following 
\textbf{Tool Packing Task}
for the described robotic system: 
"\textit{The system must autonomously pick all the tools placed one-by-one on a flat surface, transport and place them in a toolbox such that the space in the toolbox is uniformly occupied.}"

The reasons we chose this task are as follows: i) it requires a combination of a large number of perception, reasoning and motor skills; thus, a wide range of components are implemented and integrated, and ii) it is an interesting problem for any industry that can benefit from automated pick-and-place operations.

\begin{figure*}
    \centering
    \includegraphics[width=0.95\textwidth]{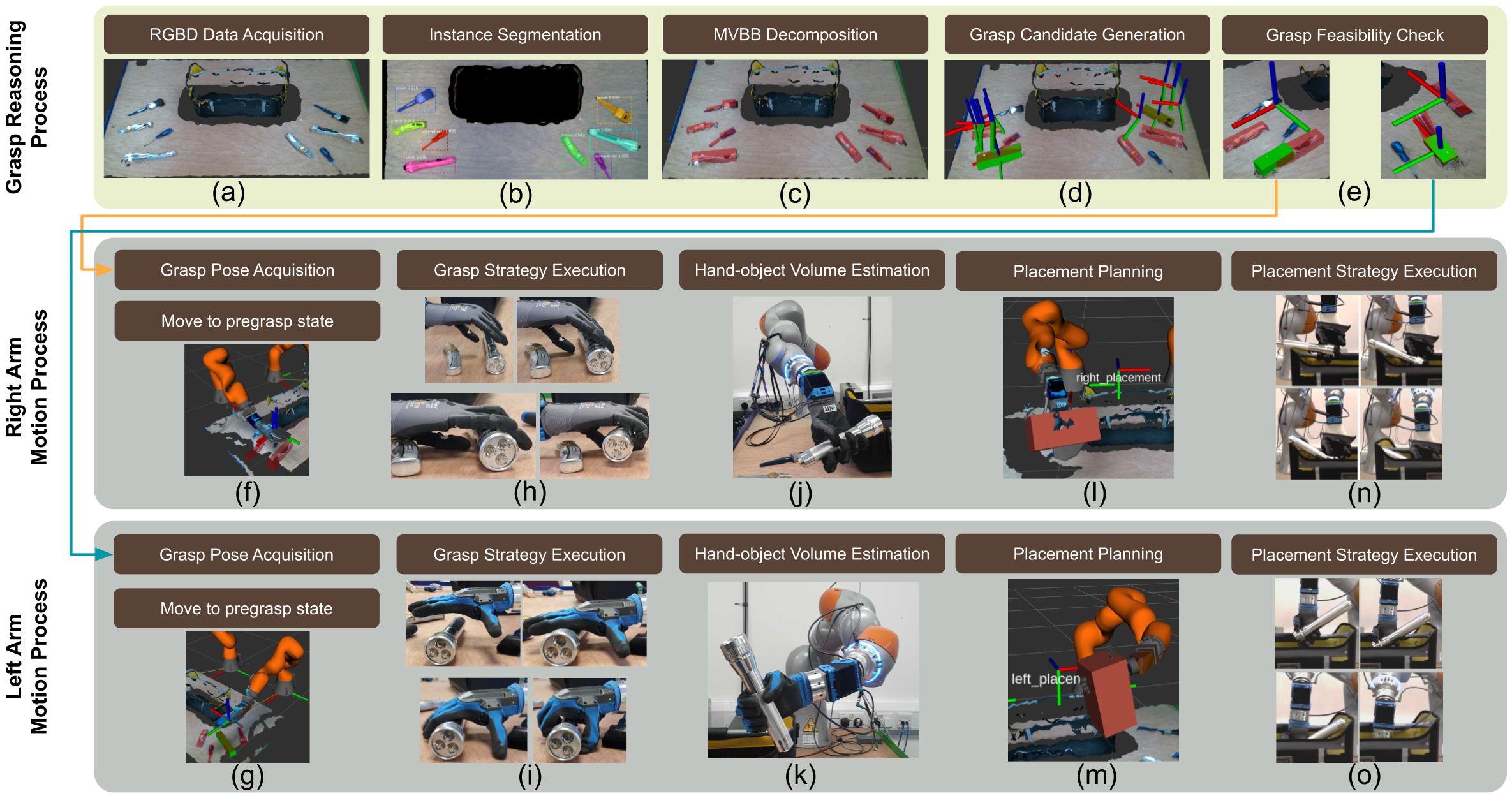}
    \caption{This diagram summarizes the main steps of the tool-packing pipeline. There are three parallel processes that run repeatedly. The grasp reasoning process converts raw RGB-D input to feasible grasp strategies for the two arms. The arms execute these strategies in parallel. The object placement is the only part executed sequentially.}
    \label{fig:tool_packing_pipeline}
\end{figure*}

\subsection{Hardware setup}
\label{system_setup}
The dual-arm manipulation system that we developed 
is illustrated in Fig. \ref{fig:bimanual_system}. 
Our system is designed as a benchmarking testbed for various research components and, as such, each of its arms is equipped with a different end-effector and force-torque (FT) sensor.
The system consists of two KUKA LBR iiwa 14 R820 arms, rigidly mounted on one end of a wooden bench and at a distance of $80cm$ from each other. 
One arm is equipped with a JR3 6-axes FT sensor and uses the ARMAR-6 Hand \cite{Asfour2018} as its end-effector that is an anthropomorphic, underactuated five-finger hand based on the
TUAT/Karlsruhe hand mechanism. It has 14 joints and is driven by 2
brushed DC motors (one motor for the thumb, one motor for
the four fingers).
The other arm has an Optoforce FT sensor and uses the qb SoftHand Research \cite{Catalano2014} as the end-effector. It is an anthropomorphic, soft and underactuated five-finger hand with 19 joints controlled by a single motor. Its fingers consist of rigid phalanges connected by elastic ligaments, thus making the hand compliant, yet robust.
Finally, an external Intel\textregistered{} RealSense\texttrademark{} D415 depth camera is mounted on the ceiling above the bench and is used to perceive the scene. All the software is running on a single Ubuntu 18.04 computer and the system has been developed using ROS melodic.

The number of hardware components in the system means that considerable integration effort is required to just launch them in isolation, and even more as parts of the same system. This effort consists of performing 
both high-level and low-level integration decisions, as it is described in Section \ref{systems_integration}.

\subsection{System functionality overview}
\label{system_overview}
The main steps of the proposed tool-packing pipeline and their respective outcomes are illustrated in Fig. \ref{fig:tool_packing_pipeline}. The pipeline consists of the following parts: 1) grasp planning (Fig. \ref{fig:tool_packing_pipeline}.a-e), 2) grasp control (Fig. \ref{fig:tool_packing_pipeline}.f-i), 3) placement planning (Fig. \ref{fig:tool_packing_pipeline}.j-m), and 4) placement execution (Fig. \ref{fig:tool_packing_pipeline}.n-o). 
As the maintenance tools in our task instance are relatively lightweight and small, we consider that a single arm is sufficient to grasp each object. 
As such, the main benefit of a dual-arm system lies in the possibility of executing two tasks in parallel.
Therefore, we configured the grasp reasoning and motion processes of the system so that the amount of time one of the two arms is idle is reduced.

The aim of an autonomous robotic pick-and-place system is to find {grasp affordances} and sequence them with {placement affordances}.
{Grasp affordances} refer to opportunities that objects present for a particular end-effector to grasp them, while {placement affordances} are options object placement within the defined environment.

\subsubsection{Grasp planning} \label{sssec:grasp_planning}
As a first step to detect grasp affordances, we segment a partial point cloud for each tool on the bench (Fig. \ref{fig:tool_packing_pipeline}.b). 
Similar to \cite{Gabellieri2020}, we approximate each of these point clouds with multiple minimum-volume bounding boxes (MVBBs) which we consider representative of the various grasp affordances of a certain tool (Fig. \ref{fig:tool_packing_pipeline}.c).
We exploit such grasp affordances to generate candidate grasps optimized for under-actuated and soft hands.
More specifically, for each MVBB, we construct candidate pre-grasp states, which correspond to the relative 6D pose between the end-effector and the MVBB (Fig. \ref{fig:tool_packing_pipeline}.d).
We check if this pose is kinematically reachable and collision-free (Fig. \ref{fig:tool_packing_pipeline}.e), and if it is not, then the closest feasible pre-grasp pose is computed.

\subsubsection{Grasp control} \label{sssec:grasp_control}

After the system reaches a pre-grasp state (Fig. \ref{fig:tool_packing_pipeline}.f and g), 
the corresponding end-effector moves towards the object until a predefined force threshold is exceeded. This signals contact with either the object or the supporting surface.

We adopt distinct grasp control strategies for both robotic hands.
The qb SoftHand is fully closed, while the arm motion is 
regulated by a joint impedance controller.
This prevents the table surface from blocking the hand fingers when they are closing (Fig. \ref{fig:tool_packing_pipeline}.l). 
The ARMAR-6 hand is closed in two phases. Initially,  the fingers are closed around half their full range to constrain the object between the thumb and the fingers, and after a small delay the hand is closed fully (Fig. \ref{fig:tool_packing_pipeline}.h). 
During these  movements, we support the grasp control strategy by controlling the wrenches exerted by the robot hand on the bench surface. 
This guarantees that the hand neither loses contact with the supporting surface, nor that there is an excessive build up of forces during closure.

The arms lift the grasped object for a pre-defined amount of time that guarantees that the tool loses contact with the supporting surface. Then, grasp success is validated using a combination of two methods: i) reading the hand encoders, and ii) using force measurements.

\subsubsection{Placement planning} \label{sssec:placement_planning}
Our placement planning pipeline consists of the following steps:
\begin{itemize}
    \item Approximate the combined hand-object volume with a MVBB (Fig. \ref{fig:tool_packing_pipeline}.j and k);
    \item Search for a candidate placement affordance and validate whether this affordance and the corresponding placement motion are feasible (Fig. \ref{fig:tool_packing_pipeline}.l and m).
\end{itemize}

Given the hand-object MVBB and a status of current toolbox occupancy, the placement affordance that leads towards a more uniform toolbox occupancy is identified. 
Since only \textit{top-down} placements are feasible in the toolbox, the placement affordances  narrow down to 6D poses for the end-effector above the toolbox. 
The hand-object MVBB is used by the motion planner to validate whether a certain placement pose and the corresponding placement motion are kinematically reachable and collision-free (i.e there are no hand-object-toolbox collisions). 
If the above is not possible, the system attempts to find the closest feasible placement pose by sampling a predefined 2D grid above the toolbox. 

\subsubsection{Placement execution} \label{sssec:placement_execution}
The two arms are prevented from simultaneously placing tools in the toolbox, due to the space constraints of the toolbox.
Therefore, an arm must acquire exclusive access to the area above the toolbox and prevent the other arm from accessing it simultaneously.

Once a feasible placement affordance is found and access to the placement area is acquired, the corresponding arm moves above the toolbox and lowers its end-effector until a contact with the bottom of the toolbox is detected and the object is released. 
The force-torque measurements of the wrist-mounted force-torque sensor are used to distinguish between a contact of the end-effector with the bottom of the toolbox and contacts with the toolbox sides. 
This can occur because of partial knowledge of the combined hand-object volume through point cloud data. 
In case the end-effector collides with the sides of the toolbox, we use the force-torque feedback to repulse the arm towards the center of the toolbox and out of collision (Fig. \ref{fig:tool_packing_pipeline}.n and o). 

\section{SYSTEMS INTEGRATION CASE-STUDY \label{case_study}}
In this section, we showcase how the methodology   proposed in Section \ref{systems_integration} is employed to achieve a smooth integration for the system presented in Section \ref{dual_arm_system}. 

\begin{table}[t!]
    \caption{Priority factor per low-level integration condition for the dual-arm system.}
    \label{priority_table}
    \begin{center}
        \begin{tabular}{|M{0.2\linewidth}||M{0.05\linewidth}|M{0.05\linewidth}|M{0.05\linewidth}|M{0.05\linewidth}|M{0.05\linewidth}
% |M{0.05\linewidth}
|}
\hline
 & \textbf{C\ref{req_interface}} & 
 \textbf{C\ref{req_syntax}} & 
 \textbf{C\ref{req_performance}} &
 \textbf{C\ref{req_quality}} &
 \textbf{C\ref{req_support}} 
%  & \textbf{C\ref{req_license}}
 \\
\hline
 \textbf{Priority factor} &
 1.0 & 
 1.0 & 
 0.5 &
 0.5 &
 0.5 
%  & 0.5
 \\
\hline
\end{tabular}
    \end{center}
\end{table}

\begin{table*}[t!]
    \caption{High-level integration effort per component of the dual-arm system.}
    \label{high_level_table}
    \begin{center}
        \begin{tabular}{|M{0.2\linewidth}||M{0.11\linewidth}|M{0.11\linewidth}|M{0.11\linewidth}|M{0.11\linewidth}||M{0.15\linewidth}|}
\hline
 & \textbf{Conflict resolution effort} & 
 \textbf{Native inclusion impact} & 
 \textbf{Blueprint creation effort} &
 \textbf{Containerisation impact} &
 \textbf{High-level integration approach}
 \\
\hline
 \textbf{ROS ecosystem} &
 Low & 
 Low & 
 Medium &
 Low &
 native inclusion
 \\
\hline
 \textbf{ARMAR-6 Hand ROS driver} &
 High & 
 High & 
 High &
 Low &
 containerisation
 \\
\hline
 \textbf{MaskRCNN} &
 Medium & 
 Medium & 
 Low &
 Low &
 containerisation
 \\
\hline
 \textbf{MoveIt (online motion planning \& control)} &
 Low & 
 Low & 
 Medium &
 Low &
 native inclusion
 \\
\hline
\end{tabular}
    \end{center}
\end{table*}

\begin{table*}[t!]
    \caption{Low-level integration effort per component of the dual-arm system.}
    \label{low_level_table}
    \begin{center}
        \begin{tabular}{|M{0.2\linewidth}||M{0.05\linewidth}|M{0.05\linewidth}|M{0.05\linewidth}|M{0.05\linewidth}|M{0.05\linewidth}
||M{0.05\linewidth}||M{0.2\linewidth}|}
\hline
 & \textbf{C\ref{req_interface}} & 
 \textbf{C\ref{req_syntax}} & 
 \textbf{C\ref{req_performance}} &
 \textbf{C\ref{req_quality}} &
 \textbf{C\ref{req_support}} &
 \textbf{$e_f$} &
 \textbf{Low-level integration approach}
 \\
\hline
 \textbf{ROS ecosystem} &
 0.0 & 
 0.0 & 
 0.0 &
 0.5 &
 0.5 &
 0.095 &
 direct integration
 \\
\hline
 \textbf{ARMAR-6 Hand ROS driver} &
 2.0 & 
 0.0 & 
 0.0 &
 0.0 &
 1.5 &
 0.333 &
 augmentation
 \\
\hline
 \textbf{MaskRCNN} &
 3.0 & 
 0.0 & 
 0.0 &
 1.0 &
 1.5 &
 0.524 &
 augmentation
 \\
\hline
 \textbf{MoveIt (online motion planning \& control)} &
 3.0 & 
 0.0 & 
 1.5 &
 0.5 &
 0.5 &
 0.571 &
 re-implementation
 \\
\hline
\end{tabular}
    \end{center}
\end{table*}

As the system is still in the research and development stage, we consider more important the low-level integration conditions C\ref{req_interface} (Interface) and C\ref{req_syntax} (Syntax) as those enable  quick development of a baseline system to iterate on. Therefore, the priority factors $p_{C_1}$ (Interface) and $p_{C_2}$ (Syntax) have been set to $1.0$ to give more impact to the corresponding conditions, as shown in Table \ref{priority_table}.
A summary of the high-level and low-level integration decisions made for the dual-arm system components is presented in Tables \ref{high_level_table} and \ref{low_level_table} respectively. 

\subsection{ROS ecosystem}
\label{ssec:ros_ecosystem}

As the dual-arm system is developed using ROS, several ROS components were included natively. It was not required to containerise any ROS-based component as ROS was the development tool used for several elements of the system, such as the KUKA IIWA arm drivers or the grasping and placement planning components. Therefore, both the high-level and low-level integration efforts of including ROS-based third-party components natively and directly were low.

\subsection{ARMAR-6 Hand ROS driver}

The ARMAR-6 Hand used by the dual-arm system is actuated by a real-time controller developed using the ArmarX robotics framework\footnote{ArmarX Framework: https://armarx.humanoids.kit.edu/index.html} by the Karlsruhe Institute of Technology. The version of ArmarX required relies on Ubuntu 14 and cannot be run directly on the robot's computer, which uses Ubuntu 18. As such, a native inclusion approach requires high effort and has significant impact on the system. On the other hand, a container blueprint which interfaces ArmarX to ROS needs to be created, which also involves high effort, but the impact of this approach on the system is low. Therefore, the high-level integration approach we pursued was containerisation.

Regarding low-level integration, we identified that there is a medium amount of effort required to create a bridge with ROS for the existing ArmarX interface (C\ref{req_interface} - Interface). Also, C\ref{req_support} (Support) is unmet because this specific component version was deprecated and the support community of ArmarX is still in its early stage. Therefore, according to Eq. \ref{integration_equation}, there is a moderate amount of effort ($e_f=0.333$) associated with the integration of this component. However, developing an alternative controller from scratch would require much more effort, so we decided to use the existing component, but augment it so that it can be interfaced with ROS.
    
\subsection{MaskRCNN}

MaskRCNN is a neural network used for object segmentation \cite{maskrcnn17}. The dual-arm system uses a customised version of MaskRCNN written in Tensorflow\footnote{https://www.tensorflow.org/}. The version of Tensorflow used requires older GPU drivers not supported officially by Ubuntu 18. Adopting such drivers would limit the rest of the system in using GPU computation. On the other hand, a container blueprints of MaskRCNN were already available in the SecondHands project's repositories and ready to be used. As such, minimal effort was required to adapt its blueprint to our system. Therefore, the high-level approach pursued for MaskRCNN was containerisation.

The custom version of MaskRCNN is a SecondHands project output which is not maintained after the end of the project (C\ref{req_support} - Support). Also, MaskRCNN does not have documentation and tests (C\ref{req_quality} - Code quality). Despite those shortcomings, it requires less effort to adapt MaskRCNN interface (C\ref{req_interface} - Interface) rather than to re-implement the component, therefore augmentation was chosen as low-level integration approach.

\subsection{Online Motion Planning}

The dual-arm system uses MoveIt \cite{moveit} to perform online motion planning. MoveIt is part of the ROS ecosystem and, as the dual-arm system is developed using ROS, there are no conflicts to resolve as already explained in Section \ref{ssec:ros_ecosystem}. As such the high-level integration approach pursued is native integration.

The ROS drivers of the robot arms are not implemented using the ROS Control framework, but this is the only way MoveIt interfaces with robot hardware (C\ref{req_interface} - Interface). Additionally, MoveIt's online motion planning functionality is not satisfactory for the given system, as the replanning is performed only if the current plan becomes infeasible, e.g. due to collision with a moving obstacle (C\ref{req_performance} - Performance). Finally, MoveIt's online planning features are not extensively documented and MoveIt does not offer official support to users (C\ref{req_quality} - Code quality, C\ref{req_support} - Support). Based on these considerations, Eq. \ref{integration_equation} provided a high score ($e_f=0.571$) and the low-level integration approach chosen was the re-implementation of MoveIt's online motion planning functionality and ROS Control's joint trajectory following capabilities.

\medskip

Based on the above considerations on the integration level required for each component, we have assigned the estimated integration effort and impact on the system, as it is shown in Table \ref{high_level_table}. The effort coefficient $e_f$ was calculated for each low-level integration condition (Table \ref{low_level_table}), and  low-level integration becomes increasingly complex as the $e_f$ coefficients increase. The choice between containerisation and native integration high-level integration approaches is guided by the estimated efforts and impact. The approach chosen aims to minimise the effort and impact of integration on the system. By employing our methodology we managed to choose the integration technique which was most appropriate for every component of the system, without committing to large unneeded integration works.

\colorlet{gray}{red!80!black}
\colorlet{yellow}{orange}
\colorlet{green}{blue}

\section{EXPERIMENTAL VALIDATION \label{experiments}}
Integration quality is inherently difficult to measure for three main reasons:
i) integration is often considered as a binary concept, i.e. it is either successful or not, ii) it is difficult to quantify which aspects of system performance are affected by system integration, and iii) integration methodologies affect not only system performance but also the amount of time required to build a system.
In this section, we address the above issues by proposing a set of system performance evaluation metrics (Section \ref{success_metrics}) that demonstrate that  the proposed methodology of Section \ref{systems_integration} leads to a well-functioning system (Section \ref{system_evaluation}). In addition, we define an experimental protocol (Section \ref{ssec:experimental_protocol}) in order to compute these metrics. Finally, we evaluate the ability of our methodology to function as a predictor of integration time (Section \ref{time_eval}).

\subsection{Experimental Protocol} \label{ssec:experimental_protocol}
The goal of the protocol is to assess the robustness, reliability and operational speed of the bi-manual manipulation system for the given packing task.
Four tools were identified as representative examples of a maintenance technician's toolset: a torch, a cutter, a brush and a screwdriver. The selected toolbox is a $55cm \times 25cm \times 28cm \: (L \times W \times H)$ Stanley Open Tote Bag.

To facilitate the evaluation of the system, the task is divided in two phases: grasping and placement. 
The grasping phase consists of grasp planning (Section \ref{sssec:grasp_planning}) and execution of grasp strategies (Section \ref{sssec:grasp_control}). 
The grasping phase ends when the tool loses contact with the supporting surface, i.e. when the end-effector supports the. full weight of the tool. 
The placement phase comprises planning (Section \ref{sssec:placement_planning}) and execution of arm motion (Section \ref{sssec:placement_execution}) from the moment when the hand grasped and lifted the tool, up to the release of the tool inside the toolbox. 
A single pick-and-place attempt is considered successful if the tool is in the toolbox after the {placement} phase.

Below, we describe the experimental procedure for a single execution of the given task.
As a first step, all four tools of the object set are placed on the bench within the workspace of each arm. 
Tools are placed randomly, but it is ensured that there is more than $3cm$ gap between them to prevent the creation of clutter.
The robot arms move to an initial pre-defined position and  autonomously pick and place the tools one-by-one in an arbitrary order until there is no pick option available. 
The system considers that there are no pick options if:
 i) there are no more tools on the bench, ii) the remaining tools are not detected by the system, or iii) the system is unable to find a feasible way of grasping any of the remaining tools. 
No external intervention is allowed during the execution, therefore objects dropped out of the robots' workspace are not reintroduced.

\subsection{Metrics for System Evaluation}
\label{success_metrics}
We benchmark the system performance over 10 consecutive executions of the task.
$\mathcal{H}$ denotes the set of robot hands used in the system, and $\mathcal{T}$ identifies the set of tools for the experiment.
For this experiment, $\mathcal{H} = \{ \textsc{qb \: hand}, \textsc{armar-6 \: hand} \}$ and $\mathcal{T} = \{\textsc{torch}, \textsc{cutter}, \textsc{brush}, \textsc{screwdriver} \}$.

As a next step, the task performance metrics is defined. It assess how successful a specific hand $h \in \mathcal{H}$ is at grasping tool $t \in \mathcal{T}$.
If $i$ denotes the trial number, then:
\begin{itemize}
    \item $n^i_{s_{h, t}} \in \{0, 1\}$ indicates whether hand $h$ managed to successfully pick-and-place tool $t$.
    \item $n^i_{a_{h, t}} \geq 0$ represents the number of pick and place attempts.
    \item $n^{i}_{u_{h, t}} \in \{0, 1\}$ indicates if tool $t$ was left on the bench at the end of a trial. 
\end{itemize}

The values from each trial can be used to define success-based metrics, which reflect how successful a hand is at grasping a specific tool.
In particular, the grasping precision is defined as $P_{h, t}$, and the attempt and success rates, are defined as $A_{h, t}$ and $R_{h,t}$, respectively. 
The precision score corresponds to the robustness of a hand's grasping and placement strategies. 
A high precision score means that the system is able to ensure that if a grasp affordance is detected, the hand-arm combination can successfully exploit it.
The attempt and success rates represent the effectiveness of a system at detecting a grasp affordance, and the effectiveness of the entire pick-and-place pipeline, respectively.
As such, for $N$ trials, these metrics can be formulated as follows:
    \begin{align}
        P_{h, t} &= \sum_{i=1}^N \frac{n^i_{s_{h, t}}}{n^i_{a_{h,t}}} \\
        A_{h, t} &= \sum_{i=1}^N \frac{n^i_{a_{h, t}}}{n^i_{a_{h, t}} + n^{i}_{u_{h, t}}} \\
        R_{h, t} &= \sum_{i=1}^N \frac{n^i_{s_{h,t}}}{n^i_{a_{h, t}} + n^{i}_{u_{h, t}}}
    \end{align}

Finally, the overall system performance is evaluated based on the mean picks per hour (MPPH) and the task completion score, $S$.
MPPH is a well-known metric in logistics for measuring both human and machine efficiency and throughput.
The task completion score corresponds to the average number of tools that were successfully picked and placed per trial, i.e.,
\begin{equation}
    S = \frac{1}{N} \sum_{i \in [1, N],\; h \in \mathcal{H},\; t \in \mathcal{T}} n^{i}_{s_{h, t}}.
\end{equation}

\subsection{System Evaluation Results}
\label{system_evaluation}
This section presents and analyses the benchmark results for our system. 
The success of a complex manipulation task  depends  both on  the  strategies  used  (grasp  planning,  robot control,  etc.) and on an effective integration of every component involved. However, a poor  integration is likely to have a visible  negative  effect  on the overall  system  performance.  Therefore,  in the following discussion we identify what affects the performance of our system and we demonstrate that our integration methodology does not cause any system  failures, but on the contrary it facilitates the development of a well-functioning system.
According to the overall system performance metrics (see Table \ref{table:results_high_level}), our proposed integration methodology  leads to a system that is able to complete more than 75\% of the given packing task on average. 
We should note that the evaluated system is still in the research stage and is not optimized for speed of operation as it would be required in production. 
As a consequence, MPPH is quite low, even in the cases where the system manages to fully complete the task, i.e. $S=8$.

\begin{table}[t]
 \caption{Overall Performance Metrics for the dual-arm system.}
 \centering
 \begin{tabular}{|c | c | c |} 
  \hline
  \textbf{Metric} & \textbf{Mean} & \textbf{Stdev}\\
  \hline
  Task completion score (max of 8) & 6.27 & 1.56 \\
  \hline
  Mean picks per hour & 37.90 & 4.25 \\ 
  \hline
 \end{tabular}
 \label{table:results_high_level}
\end{table}

In order to facilitate system introspection and identify potential integration issues, we monitor system failures during task execution. More specifically, we identified the following failure types:

\begin{itemize}
    \item Visual detection failure - false positive and false negative object detections by the MaskRCNN model.
    \item Grasping strategy failure - any other type of failure that occurs during the {grasping} phase of the pick-and-place pipeline. For example, the hand might be unable to cage a tool or a grasp is not robust enough to lift the tool from the bench. 
    \item Placement strategy failure - errors occurring during the {placement} phase. For instance, the robots might drop tools outside of the toolbox, because of incorrect decisions on the placement location or erroneous toolbox pose detection.
\end{itemize}

\begin{figure}[!htbp]
    \centering
    \includegraphics[width=0.45\textwidth]{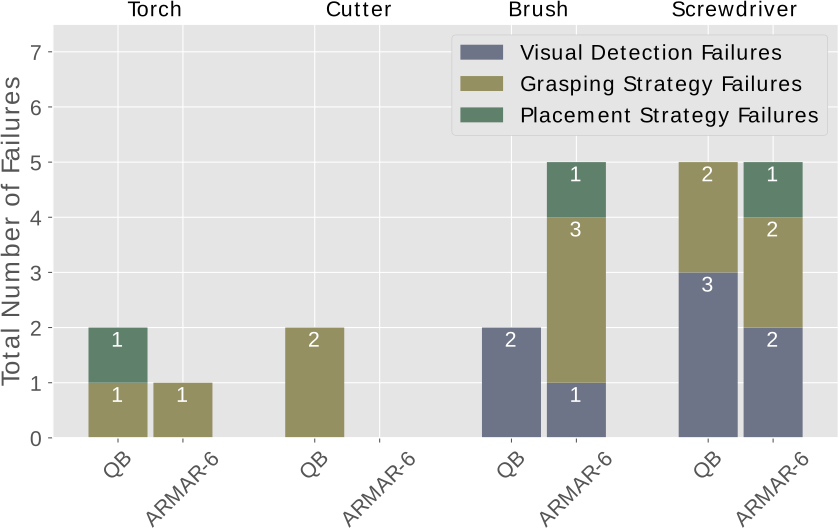}
    \caption{Number of system failures per hand and tool.}
    \label{fig:results_failure_modes}
\end{figure}

\begin{figure}[!htbp]
    \centering
    \includegraphics[width=0.45\textwidth]{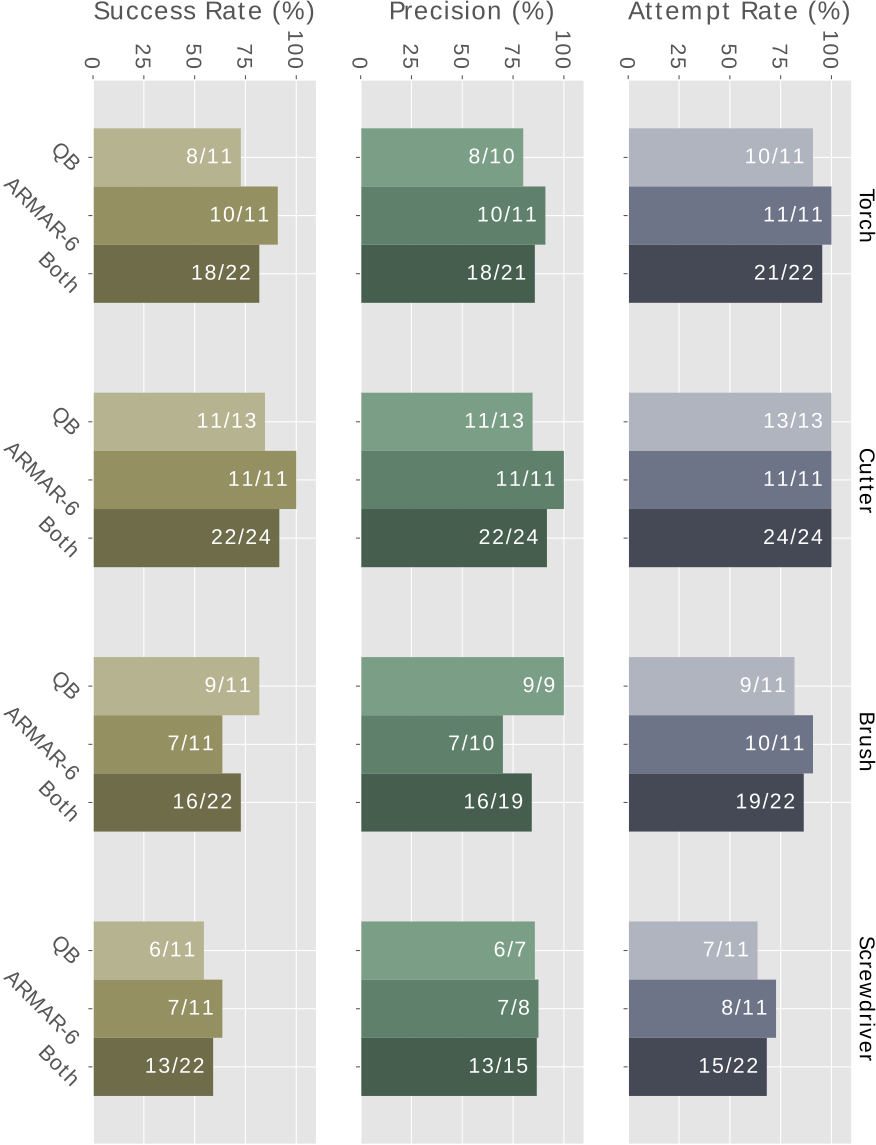}
    \caption{Success-based performance metrics per tool.}
    \label{fig:results_success_rates}
\end{figure}

The number of times each of the above failures occurs per hand and object is presented in Fig. \ref{fig:results_failure_modes}, while the corresponding success-based performance metrics are illustrated in Fig. \ref{fig:results_success_rates}.
Both figures highlight the effect that object geometry has on system performance. 
More specifically, it is apparent that big objects like torches and cutters are always detected by the visual system.
This is not always the case for smaller tools like brushes and screwdrivers. 
This is because bigger objects are more likely to have features that persist through multiple pooling layers of the deep neural network model, whereas smaller visual features might be occasionally skipped. 
This observation is also confirmed by the fact that attempt rates for torches and cutters are higher compared to brushes and screwdrivers (see Fig. \ref{fig:results_success_rates}). 
Regarding grasping strategy failures, we observe that grasp success depends on hand morphology and, more specifically, on the ability of a hand to restrict object displacement while closing. As a consequence, the qb SoftHand Research is performing better for grasping very flat tools like the brush, as confirmed by the precision and success rate metrics for brushes in Fig. \ref{fig:results_success_rates}.
Finally, a few placement failures occur for long objects due to unwanted object-toolbox collisions which are not mitigated by the force-repulsion component (see Section \ref{sssec:placement_execution}).

Overall, the experimental results verify that the proposed methodology leads to a well-integrated system. More specifically, the absence of unexpected system errors and crashes signifies successful high-level integration. Regarding low-level integration, all system components were able to interoperate with each other as needed and performed well enough given the development stage of the evaluated system and the challenging nature of the task.

\subsection{Integration Time Prediction Evaluation}
\label{time_eval}
This section evaluates whether our integration methodology can predict the integration time. Table \ref{timetable} depicts the expected and actual integration time for the components discussed in Section \ref{case_study}. The expected integration time was converted from the efforts of Tables \ref{high_level_table} and \ref{low_level_table} using Table \ref{time_table} and Equation \ref{integration_time_equation}. The actual integration time was derived from the logs of the Agile management software used (Jira\footnote{Jira: https://www.atlassian.com/software/jira}); the logs detail the start and completion date of each integration task. The time reported refers exclusively to the integration of the components, ignoring the implementation time of the algorithms. Overall our integration methodology predicted the integration time reliably.

\begin{table*}[t!]
    \caption{Predicted vs Actual integration time.}
    \label{timetable}
    \begin{center}
        \begin{tabular}{|M{0.11\linewidth}||M{0.11\linewidth}|M{0.11\linewidth}||M{0.1\linewidth}|M{0.11\linewidth}||M{0.13\linewidth}||M{0.11\linewidth}|}
\hline
 &
 \textbf{High-level integration effort} &
 \textbf{Expected high-level integration time} & 
 \textbf{Low-level integration effort coefficient $e_f$} &
 \textbf{Expected low-level integration time} & 
 \textbf{Expected total integration time} &
 \textbf{Actual integration time}
 \\
\hline
 \textbf{ROS ecosystem} &
 \textbf{Low} conflict resolution effort + \textbf{Low} native inclusion impact&
 0-4 person days& 
 0.095 &
 $\sim$ 1 person days & 
 $\sim$ 1-5 person days &
 $\sim$ 1 person days
 \\
\hline
 \textbf{ARMAR-6 Hand ROS driver} &
 \textbf{High} blueprint creation effort + \textbf{Low} containerisation impact &
 $>$6 person days & 
 0.333 &
 $\sim$ 3 person days & 
 $>$9 person days &
 10 person days
 \\
\hline
 \textbf{MaskRCNN} &
 \textbf{Low} blueprint creation effort + \textbf{Low} containerisation impact &
 0-4 person days & 
 0.524 &
 $\sim$ 5 person days & 
 $\sim$ 5-9 person days &
 4 person days
 \\
\hline
 \textbf{MoveIt (online motion planning \& control)} &
 \textbf{Low} conflict resolution effort + \textbf{Low} native inclusion impact &
 0-4 person days & 
 0.571 &
 $\sim$ 6 person days & 
 $\sim$ 6-10 person days &
 7 person days
 \\
\hline
\end{tabular}
        % Swap the above with predicted_vs_actual to see the other option
    \end{center}
\end{table*}
\section{Discussion\label{discussion}}

In this article, we propose an integration methodology, summarised by the workflow in Fig. \ref{fig:integration_flowchart}, that assesses the effort and impact of integrating a robotic component in a complex system. 
We showcase the application of our approach to an industrial case-study: a dual-arm manipulation system that consists of a diverse set of research components. We use the system to demonstrate how to practically apply our methodology and address real-world integration problems. By following the flowchart (Fig. \ref{fig:integration_flowchart}) it is possible to understand which requirements are key to facilitate the adoption of academic research into industrial systems. 
We evaluate the performance of the system  with a benchmark, designed to measure the robustness, reliability and operational speed of the whole system as well as the relative effectiveness of the two employed end-effectors when integrated within the same system. Our integration methodology was employed alongside the Agile Scrum \cite{lei2017} development methodology since it helps estimating the time required for the integration of a component. This is a key factor in every Agile methodology, where time is expected to be correctly estimated by the developers. Nevertheless, our methodology is agnostic of the development methodology adopted by a team.

System integration is a topic rarely addressed by the robotics research community and considered an overhead rather than an important part of a system. However, a robotic system needs to be integrated in a coherent structure to be applied to a real-world use case. Additionally, valuable scientific contributions might be lost if it is unfeasible to integrate them. For a successful translation of robotic research into industrial applications, the research community needs to apply established integration methods. Moreover, the integration methodology has to be grounded by the development of real-world robotic systems of high technology readiness level. Therefore, methods of and learning from integration of complex robotic systems need to be shared among the robotics community. Indeed, system complexity poses an integration challenge which cannot be replicated in a scenario prepared to study integration as an isolated factor. Real-world robotic systems face integration problems difficult to anticipate; their resolution, especially if performed systematically, provides knowledge applicable to the integration of other systems. Such knowledge cannot be obtained from the integration of simple systems. Hence, sharing integration solutions of complex systems would allow to establish good engineering practices and to avoid common shortcomings in designing and integrating robotic systems. 

\section{Conclusion\label{conclusion}}

This paper is a description of our proposed integration methodology and an example of how a complex system can be integrated together in an efficient manner. We believe that our work can help the robotics research community in deciding which integration approach to follow and can be used as a reference for integrating systems of similar complexity. Given the complexity of the described system, developed to solve an industrial problem, it is not viable to repeat the integration using a second methodology. Therefore, as future work, we will apply and re-evaluate our integration methodology on different complex robotic systems and compare the new integration experience with the one presented in this work.

\bibliographystyle{IEEEtran}
%\bibliography{IEEEabrv,references}

\end{document}